\documentclass{article}

\usepackage[table]{xcolor}
\usepackage{microtype}
\usepackage{graphicx}
\usepackage{subfigure}
\usepackage{booktabs} 

\usepackage{hyperref}

\usepackage[accepted]{icml2024}

\usepackage{amsmath}
\usepackage{amssymb}
\usepackage{mathtools}
\usepackage{amsthm}

\usepackage[capitalize,noabbrev]{cleveref}

\usepackage{multirow}
\usepackage{makecell}
\usepackage{array}
\newcolumntype{P}[1]{>{\centering\arraybackslash}p{#1}}

\theoremstyle{plain}

\theoremstyle{definition}

\theoremstyle{remark}

\usepackage[textsize=tiny]{todonotes}

\icmltitlerunning{EDoRA: Efficient Weight-Decomposed Low-Rank Adaptation via Singular Value Decomposition}

\begin{document}

\twocolumn[
\icmltitle{EDoRA: Efficient Weight-Decomposed Low-Rank Adaptation\\ via Singular Value Decomposition}



\icmlsetsymbol{equal}{*}

\begin{icmlauthorlist}
\icmlauthor{Hamid Nasiri}{lanc}
\icmlauthor{Peter Garraghan}{lanc}
\end{icmlauthorlist}

\icmlaffiliation{lanc}{School of Computing and Communications, Lancaster University, Lancaster, UK}

\icmlcorrespondingauthor{Hamid Nasiri}{h.nasiri@lancaster.ac.uk}
\icmlcorrespondingauthor{Peter Garraghan}{p.garraghan@lancaster.ac.uk}

\icmlkeywords{Machine Learning, ICML}

\vskip 0.3in
]



\printAffiliationsAndNotice{}  

\begin{abstract}
Parameter-efficient fine-tuning methods, such as LoRA, reduces the number of trainable parameters. However, they often suffer from scalability issues and differences between their learning pattern and full fine-tuning. To overcome these limitations, we propose Efficient Weight-Decomposed Low-Rank Adaptation (EDoRA): a novel PEFT method that decomposes pre-trained weights into magnitude and directional components. By freezing low-rank matrices, initializing them by singular value decomposition, and introducing a small trainable matrix between them, EDoRA achieves substantial reduction in trainable parameters while maintaining learning capacity. Experimental results on the GLUE benchmark demonstrate that EDoRA achieves competitive or superior performance compared to state-of-the-art methods, such as LoRA and DoRA, with up to 30x fewer trainable parameters. This makes EDoRA a highly efficient solution for adapting LLMs to diverse tasks under memory-constrained settings. Code is available at \href{https://github.com/Hamid-Nasiri/EDoRA}{https://github.com/Hamid-Nasiri/EDoRA}.
\end{abstract}
\section{Introduction}
\label{introduction}

Large Language Models (LLMs) have shown remarkable success in a variety of applications, such as Natural Language Processing \citep{thirunavukarasu2023large,min2023recent} and multi-modal tasks \citep{liu2023medical,cui2024survey,jin2024chat}. However, fine-tuning these models for specific downstream tasks can be computationally expensive. Parameter-efficient fine-tuning (PEFT) methods address this challenge by only updating a small subset of the model’s parameters. Low-Rank Adaptation (LoRA) \citep{hu2021lora} is one of the most popular PEFT methods due to its good generalization, simplicity, and efficiency. Moreover, it does not add additional latency during the inference since it can merge with pre-trained weights before inference.

LoRA efficiently reduces the number of trainable parameters. Inspired by LoRA, researchers \citep{edalati2022krona,fawi2024curlora,zhang2023adalora,balazy2024lora,kopiczko2023vera} have developed adaptation methods to decrease the number of parameters further while enhancing performance. Several studies \cite{zhang2023adalora,balazy2024lora, gu2024sara} applied various matrix decomposition methods such as CUR matrix decomposition \cite{fawi2024curlora} and Singular Value Decomposition (SVD) \cite{zhang2023adalora,balazy2024lora, gu2024sara} to reduce LoRA's trainable parameters. While most methods focused on decreasing the number of trainable parameters, others \cite{liu2024dora,kalajdzievski2023rank} have concentrated on improving LoRA’s learning pattern and stability so that it more closely resembles full fine-tuning.


Although these methods effectively reduce the number of trainable parameters in LoRA, they still need substantial storage and computational resources, especially in scenarios requiring large-scale personalized or task-specific adaptation \citep{balazy2024lora}. Existing methods also suffer from a scaling problem: their parameter count increases with model dimensionality. This dependency can substantially increase trainable parameters, particularly for LLMs with higher hidden dimensions. Increasing the number of parameters can exacerbate risks in overfitting, especially when adapting to smaller downstream datasets \citep{qin2024bidora}.

Beyond these scaling issues, LoRA faces another key challenge with its learning pattern differs significantly from that of full fine-tuning. This difference potentially limits LoRA’s learning capacity. LoRA tends to update both the magnitude and direction of weights proportionally, exhibiting a strong positive correlation between the changes in these two components. This contrasts with the learning pattern observed in full fine-tuning \citep{liu2024dora}. LoRA's tendency to change the magnitude and direction proportionally, can lead to unnecessary adjustments to the pre-trained weights, potentially affecting the model's performance. To address these issues, we require a technique that can replicate full fine-tuning learning pattern and mitigate the scalability problem by reducing the number of trainable parameters.

In this paper we present an Efficient Weight-Decomposed Low-Rank Adaptation (EDoRA). EDoRA employs a decomposition strategy to decompose pre-trained weights into magnitude and directional components, freezes low-rank matrices, and introduces a small trainable matrix between them. This design replicates the learning dynamics of full fine-tuning, mitigating scalability issues and reducing the risk of overfitting. EDoRA also uses SVD for its initialization phase, which helps ensure that the adaptation process starts from a subspace aligned with the most important features of the pre-trained model. 

The major contributions of this paper are as follows:


\begin{itemize}
    \item We introduce EDoRA, a highly parameter-efficient PEFT approach that leverages a decomposition strategy and SVD-based initialization to overcome the scalability and learning pattern limitations of existing methods. 
    \item We evaluate EDoRA's performance on the GLUE benchmark. In terms of average performance over six different tasks, EDoRA outperforms LoRA and other state-of-the-art methods, such as DoRA.
    \item We present a parameter efficiency analysis of EDoRA compared to LoRA and DoRA. The results highlight that EDoRA, on average, reduces the trainable parameters by over 45x compared to LoRA and DoRA when applied to the GPT-3 model.
\end{itemize}

\section{Related Work}
\label{related_work}

Fine-tuning LLMs is often extremely costly due to their size. PEFT methods tackle this challenge by adapting large models for downstream tasks by training a small number of parameters. These parameters can be a subset of the model’s existing parameters or entirely new ones added to the model \citep{lialin2023scaling}. This significantly reduces both computational and storage costs. PEFT techniques fall into three main categories: 1. Additive Methods, 2. Selective Methods and 3. Reparametrization-based Methods. 

\subsection{Additive Methods}
The core concept of additive methods is to enhance the existing pre-trained models by introducing additional parameters or layers to the original frozen backbone. Additive methods include two main subcategories: Adapter-based methods and Prompt-based methods. Adapters, as presented by Houlsby et al. \yrcite{houlsby2019parameter}, add small fully-connected networks after transformer sub-layers \citep{lialin2023scaling}. They produce a compact and flexible model by introducing only a small number of trainable parameters for each task, allowing new tasks to be added without the need to revisit earlier ones. AdapterHub \citep{pfeiffer2020adapterhub}, a framework for adapting transformers, was developed by Pfeiffer et al. It allows dynamic “stitching-in” of pre-trained adapters for various tasks. 

The second subcategory of additive methods is Prompt-based approaches. The main idea behind these methods is to introduce additional soft tokens to the initial input and concentrate exclusively on fine-tuning these trainable vectors. Vu et al. \yrcite{vu2021spot} proposed SPOT (Soft Prompt Transfer). SPOT initially learns a prompt on one or more source tasks and then uses it to initialize the prompt for a target task. The author showed that SPOT substantially improves the performance of prompt-tuning across various tasks; however, prompt-based methods are often hindered by their sensitivity to initialization, impacting their performance. The additive methods, whether modifying the model’s input or architecture, lead to higher inference latency compared to the baseline model \citep{liu2024dora}.

\subsection{Selective Methods}
Selective methods fine-tune only a selection of layers or parameters of the model. One subcategory of selective methods is sparse update techniques, which can disregard the model’s overall structure and selectively update individual parameters. Sung et al. \yrcite{sung2021training} developed the FISH (Fisher Induced Sparse Unchanging) method. FISH applies a fixed sparse mask to the model’s parameters, selecting a subset for updating over multiple iterations. It creates the mask using the top $n$ parameters with the highest Fisher information as a simple approximation of which parameters are most important for the given task. Although sparse update methods are efficient regarding memory consumption and communication overhead, their unrestricted unstructured sparsity makes them impractical on modern hardware.

\subsection{Reparametrization-based Methods}
Reparametrization-based techniques use low-rank representation to reduce the number of trainable parameters \citep{lialin2023scaling}. Unlike additive methods, these methods do not add additional computational overhead or latency during inference. Aghajanyan et al. \yrcite{aghajanyan2020intrinsic} showed that fine-tuning is efficient in low-rank subspaces. Moreover, the authors demonstrated that the subspace size that needs adaptation decreases as model size or pre-training duration increases. Low-Rank Adaptation (LoRA) \citep{hu2021lora} is the most widely known reparametrization-based method. It uses a simple low-rank matrix decomposition to parametrize the weight update and can merge with pre-trained weights before inference. The authors in \citep{hu2021lora} showed that its low-rank weight updates are highly correlated with the pre-trained weights, indicating that LoRA amplifies specific directions already present in the model’s weight space. 

Inspired by this finding, many researchers \citep{edalati2022krona,fawi2024curlora,zhang2023adalora,balazy2024lora,kopiczko2023vera} proposed adaptation methods to decrease LoRA’s trainable parameters. Edalati et al. \citep{edalati2022krona} developed KronA, which replaces LoRA’s matrix factorization with one based on the Kronecker product. This results in a better trade-off between rank and parameter count compared to LoRA, as the Kronecker product preserves the rank of the original matrices being multiplied. CURLoRA \citep{fawi2024curlora}, introduced by Fawi, applies CUR matrix decomposition to the pre-trained weight matrices instead of random initialization for the low-rank matrices in LoRA. This method leverages inverted probabilities for column and row selection during decomposition, which acts as an implicit regularization technique. By initializing the U matrix as a zero matrix and only fine-tuning it, CURLoRA aims to mitigate catastrophic forgetting and maintain model stability during continual learning. Some researchers have used other decomposition methods, such as SVD, to reduce the model’s trainable parameters.

Zhang et. al \citep{zhang2023adalora} introduced AdaLoRA in 2023. AdaLoRA is a PEFT method designed to allocate the parameter budget adaptively during the fine-tuning process. Instead of uniformly distributing the budget of incremental updates across all pre-trained weight matrices, AdaLoRA assigns a parameter budget to each weight matrix based on an importance score. AdaLoRA parameterizes incremental updates using SVD, which enables efficient pruning of singular values associated with less important updates. LoRA-XS \citep{balazy2024lora} is a highly parameter-efficient fine-tuning method. It applies SVD to the pre-trained weight matrix and uses top singular vectors to create frozen low-rank matrices, denoted as A and B. These matrices remain frozen during training. LoRA-XS inserts a small trainable matrix between A and B. Since this matrix is the only trainable component in LoRA-XS, the number of trainable parameters is significantly reduced. 

While all these methods effectively reduce LoRA’s trainable parameters, they fail to address the problem of LoRA’s tendency to update both the magnitude and direction of weights proportionally, potentially hindering its learning capacity. LoRA’s learning pattern differs significantly from full fine-tuning, which typically involves substantial adjustments in either magnitude or direction, with only minor modifications to the other \citep{liu2024dora}. To tackle this problem, Liu et al. \citep{liu2024dora} introduced DoRA. DoRA decomposes the pre-trained weight matrix into magnitude and direction components. This decomposition is inspired by Weight Normalization \citep{salimans2016weight}, which accelerates convergence by improving gradient conditioning through weight reparameterization. DoRA applies LoRA specifically to the direction component, efficiently reducing the number of trainable parameters while maintaining the advantages of low-rank adaptation. Initializing both components with pre-trained weights helps DoRA avoid initialization sensitivity, unlike Weight Normalization, which trains both components from scratch. By employing this strategy, DoRA enhances the performance and training stability of LoRA without introducing additional inference overhead.

Although DoRA improves LoRA’s learning capacity, its parameter count scales with the model’s dimensionality since the magnitude component in DoRA is an n-dimensional trainable vector, where n represents the number of columns of the weight matrix. Therefore, as the model’s hidden dimension increases, the size of the magnitude vector also increases, leading to a substantial increase in trainable parameters. This can exacerbate the risk of overfitting. Inspired by DoRA, our proposed method attempts to fill this gap by using SVD to reduce the number of trainable parameters drastically.

\section{Proposed Method}
\label{method}

\begin{figure}[htbp]
  \centering
    \includegraphics[width=0.53\textwidth]{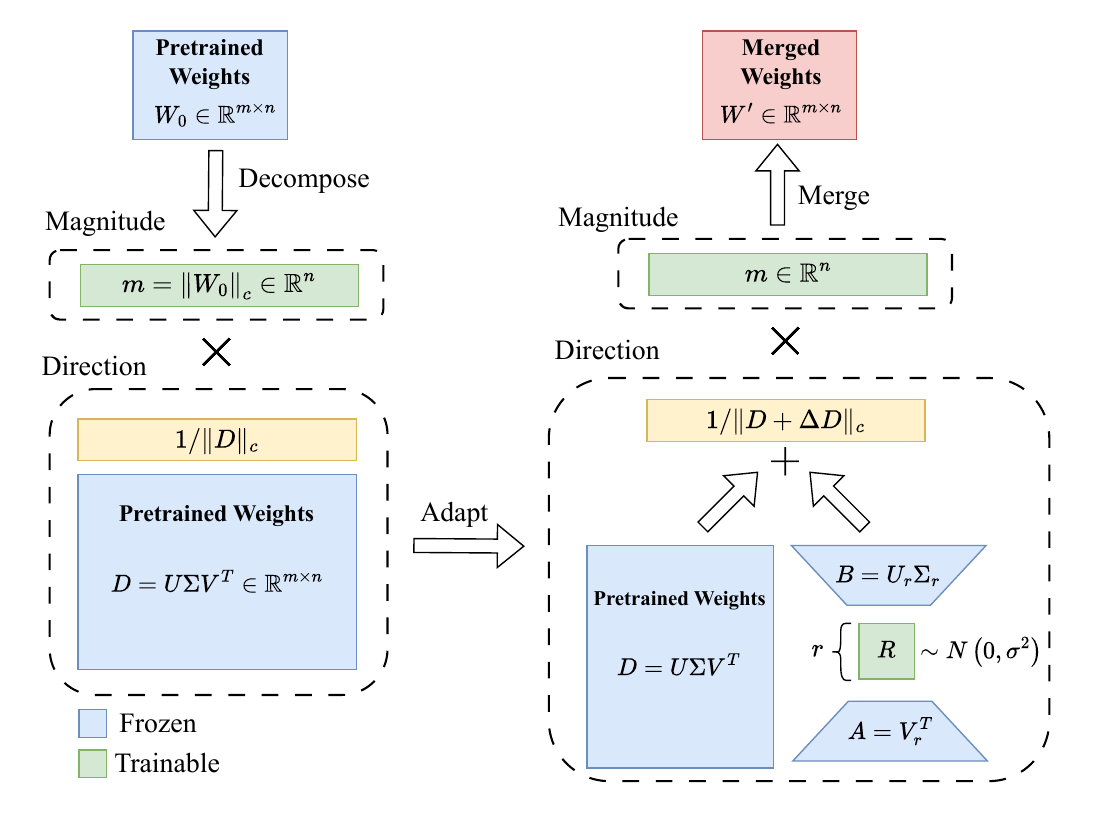}
    \caption{An overview of EDoRA}
    \label{fig:edora}
\end{figure}

The main idea of EDoRA is to decompose pre-trained weight into its magnitude and directional components. EDoRA keeps the magnitude vector trainable and, due to the substantial size of the directional component in terms of parameters, fixes low-rank matrices $A \in \mathbb{R}^{r \times n}$ and $B \in \mathbb{R}^{m \times r}$, and introduces a small trainable matrix $R \in \mathbb{R}^{r \times r}$ between them (\cref{fig:edora}).

The fine-tuned weight $W^{\prime}$ can be updated incrementally by a low-rank decomposition:

\begin{equation}
W^{\prime}=W_0+\Delta W=W_0+B A
    \label{eq:update_rule}
\end{equation}

where $W_0 \in \mathbb{R}^{m \times n}$ represents the pre-trained weight matrix, $\Delta W \in \mathbb{R}^{m \times n}$ is the low-rank weight update, $B \in \mathbb{R}^{m \times r}, A \in \mathbb{R}^{r \times n}
$, and $r \ll \min (m, n)$. During the fine-tuning process, $W_0$ is kept frozen and does not receive gradient updates, while $A$ and $B$ contain trainable parameters. $A$ is initialized with a random Gaussian distribution and $B$ is initialized with zeros, so $\Delta W=B A$ is zero at the start of training to ensure that $W^{\prime}$ equals $W_0$ before the fine-tuning.

The weight decomposition of $W \in \mathbb{R}^{m \times n}$ can be represented as:

\begin{equation}
W=m \frac{D}{\|D\|_c}
    \label{eq:w_decompose}
\end{equation}

where $m \in \mathbb{R}^n$ denotes the magnitude vector, $D \in \mathbb{R}^{m \times n}$ represents the directional matrix, and $\|\cdot\|_c$  is the vector-wise norm of a matrix across each column. Notably, each column of $D /\|D\|_c$ is a unit vector whose magnitude is specified by the corresponding entry in $m$.

Derived from Eqs. (\ref{eq:update_rule}) and (\ref{eq:w_decompose}), DoRA can be formulated as:

\begin{equation}
W^{\prime}=m \frac{D+\Delta D}{\|D+\Delta D\|_c}=m \frac{W_0+B A}{\left\|W_0+B A\right\|_c}
    \label{eq:DoRA}
\end{equation}

where $A$ and $B$ denote trainable parameters and $\Delta D$ is the direction update computed by the product of $A$ and $B$. The DoRA approach applies LoRA to the directional matrix. Moreover, the magnitude vector also receives gradient updates.

EDoRA reduces the number of trainable parameters drastically by making matrices A and B frozen and adding a small trainable matrix $R \in \mathbb{R}^{r \times r}$ to Eq. (\ref{eq:DoRA}) as follows:

\begin{equation}
W^{\prime}=m \frac{D+\Delta D}{\|D+\Delta D\|_c}=m \frac{W_0+B R A}{\left\|W_0+B R A\right\|_c}
\end{equation}

During the training, matrix $R$ is initialized with a Gaussian distribution, $N\left(0, \sigma^2\right)$, where $\sigma$ is set to a small value. The matrices $A$ and $B$ are initialized using the truncated SVD of the directional matrix $D$. Mathematically speaking, the SVD of the directional matrix can be computed as follows:

\begin{equation}
D=U \Sigma V^T
\end{equation}

where $U \in \mathbb{R}^{m \times m}, \Sigma \in \mathbb{R}^{m \times n}$, and $V \in \mathbb{R}^{n \times n}$. Then, by considering the top $r$ singular values of the directional matrix, we set $A$ and $B$ as follows:

\begin{equation}
\begin{gathered}
A=V_r^T \\
B=U_r \Sigma_r 
\end{gathered}
\end{equation}

where $U_r \in \mathbb{R}^{m \times r}$ and $V_r \in \mathbb{R}^{n \times r}$ denote left and right singular vectors corresponding to the top r singular values, respectively, and $\Sigma_r$ represents a diagonal matrix that contains the top $r$ singular values of $\Sigma$. Initializing $A$, $B$, and $R$ as mentioned, ensures that the learning process starts with a model nearly identical to the pre-trained model. 

EDoRA does not change the model architecture. It can be merged with the pre-trained weight before inference, so it does not add additional computational overhead or latency during inference. Limiting the algorithm to focus exclusively on directional adaptation while keeping the magnitude vector trainable, simplifies the task compared to the original approach. Furthermore, adding the trainable matrix $R$ guarantees that the model functions within a subspace, capturing the pre-trained weights' most important components or directions.

\subsection{Parameter Efficiency Analysis}
In this section, we evaluate the parameter efficiency of EDoRA in comparison to LoRA and DoRA. To simplify the analysis, consider a transformer model with the weight matrix $W \in \mathbb{R}^{n \times n}$. The ratio of trainable parameters of LoRA to EDoRA can be computed as follows:

\begin{equation}
\frac{P_{\text{LoRA}}}{P_{\text{EDoRA}}}=\frac{2 n r}{n+r^2}
\end{equation}

where $r$ represents the rank of low-rank decomposition. Similarly, the ratio of trainable parameters of DoRA to EDoRA can be computed as follows:

\begin{equation}
\frac{P_{\text{DoRA}}}{P_{\text{EDoRA}}}=\frac{n+2 n r}{n+r^2}
\end{equation}

Since $r \ll n$ and the model dimension $n$ grows significantly larger than the rank $r$, the advantages of EDoRA over LoRA and DoRA become increasingly evident. To illustrate this, we calculated the ratios for different LLMs and different values of $r$, as given in Table \ref{tab:parameter_efficiency}. The LLMs we considered were BERT \cite{devlin2018bert},
RoBERTa \cite{liu2019roberta},
ALBERT \cite{lan2019albert},
OPT 6.7B \cite{zhang2022opt},
GPT-3 \cite{brown2020language} and
PaLM 540B \cite{chowdhery2023palm}.

\begin{table}[htbp]
\caption{Parameter efficiency of EDoRA compared to LoRA and DoRA across various model sizes.}
\label{tab:parameter_efficiency}
\begin{tabular}{l@{\hskip 5pt}c|@{\hskip 3pt}c@{\hskip 3pt}c@{\hskip 3pt}|c@{\hskip 3pt}c}

\toprule \rule{0pt}{12pt} \multirow{2}{*}{Model} & \multirow{2}{*}{$n$}& \multicolumn{2}{c|}{$\frac{P_{\text {LoRA}}}{P_{\text {EDoRA}}}$} &  \multicolumn{2}{c}{$\frac{P_{\text {DoRA}}}{P_{\text {EDoRA}}}$} \\
 &  & $r=16$ & $r=32$ & $r=16$ & $r=32$ \\
\midrule BERT\,$_{\text{base}}$ & 768 & 24.00 & 27.43 & 24.75 & 27.86 \\
RoBERTa\,$_{\text{large}}$ & 1024 & 25.60 & 32.00 & 26.40 & 32.50 \\
ALBERT\,$_{\text{xlarge}}$ & 2048 & 28.44 & 42.67 & 29.33 & 43.33 \\
OPT 6.7B & 4096 & 30.12 & 51.20 & 31.06 & 52.00 \\
GPT-3 & 12288 & 31.35 & 59.08 & 32.33 & 60.00 \\
PaLM 540B & 18432 & 31.56 & 60.63 & 32.55 & 61.58 \\
\bottomrule
\end{tabular}
\end{table}
\section{Experiments}
\label{experiments}

This section evaluates the performance of EDoRA on the GLUE benchmark \cite{wang2018glue}. We compared EDoRA with three PEFT methods including DoRA \cite{liu2024dora}, LoRA \cite{hu2021lora} and LoRA-XS \cite{balazy2024lora} using various tasks, including inference tasks (QNLI and RTE), similarity and paraphrase tasks (MRPC and STS-B), and single-sentence tasks (CoLA and SST-2). 
\subsection{Experimental Setup}
We used the RoBERTa-base model \cite{liu2019roberta} for all experiments. To assess the effect of rank parameter on the performance, we tested various ranks ranging from $r=4$ to $r=32$ for each method. We integrated EDoRA modules into the RoBERTa model's Query, Value, Attention Output and first Fully Connected weight matrices. Hyperparameters were optimized through grid search, and the selected values are presented in Table \ref{tab:hyperparameter}. We utilized two NVIDIA V100 GPUs with 32GB of memory for training.

\subsection{GLUE Benchmark}
The performance of EDoRA and other PEFT methods (LoRA, DoRA, LoRA-XS) applied to the RoBERTa-base model is shown in Table \ref{tab:performance_glue}. We used Matthew's correlation for CoLA, Pearson correlation for STS-B and accuracy for the other tasks as evaluation metrics. To ensure robustness, we conducted five independent experiments with different random seeds. The reported results represent the median and standard deviation of these runs. 

Table \ref{tab:performance_glue} demonstrates the advantages of EDoRA over other PEFT methods across most GLUE tasks. Our approach obtained 2.53\%, 0.45\%, 2.44\% and 0.23\% higher accuracy than the second-best method in RTE, STSB, COLA and SST2 tasks, respectively. This is despite our approach having, on average, 30x fewer trainable parameters than LoRA and 32x fewer than DoRA. In MRPC and QNLI, LoRA outperformed other methods. EDoRA achieved competitive results in these two tasks compared to LoRA and DoRA while requiring significantly fewer trainable parameters. EDoRA achieved 0.49\% and 1.03\% lower accuracy than LoRA within the MRPC and QNLI tasks, respectively. EDoRA's good performance can be attributed to its decomposition strategy—which causes it to follow a learning dynamics closer to full fine-tuning—and its use of SVD for initialization. This helps EDoRA to have a more efficient adaptation process.

Moreover, EDoRA outperformed other methods on RTE and CoLA tasks, as reflected in Table \ref{tab:performance_glue}. It exceeds LoRA’s performance by 2.44\% on the COLA task and surpasses DoRA by 2.53\% on the RTE task. These tasks are particularly challenging due to their limited training data, which increases the risk of overfitting, especially when the number of trainable parameters is large. EDoRA mitigates this risk through its compact parameterization as it fine-tunes fewer parameters compared to LoRA and DoRA. It constrains the adaptation space via a small square trainable matrix and avoids unnecessary updates to pre-trained weights. This ensures better generalization on small datasets while preserving task-relevant knowledge.

\begin{figure}[t]
  \centering
    \includegraphics[width=0.46\textwidth]{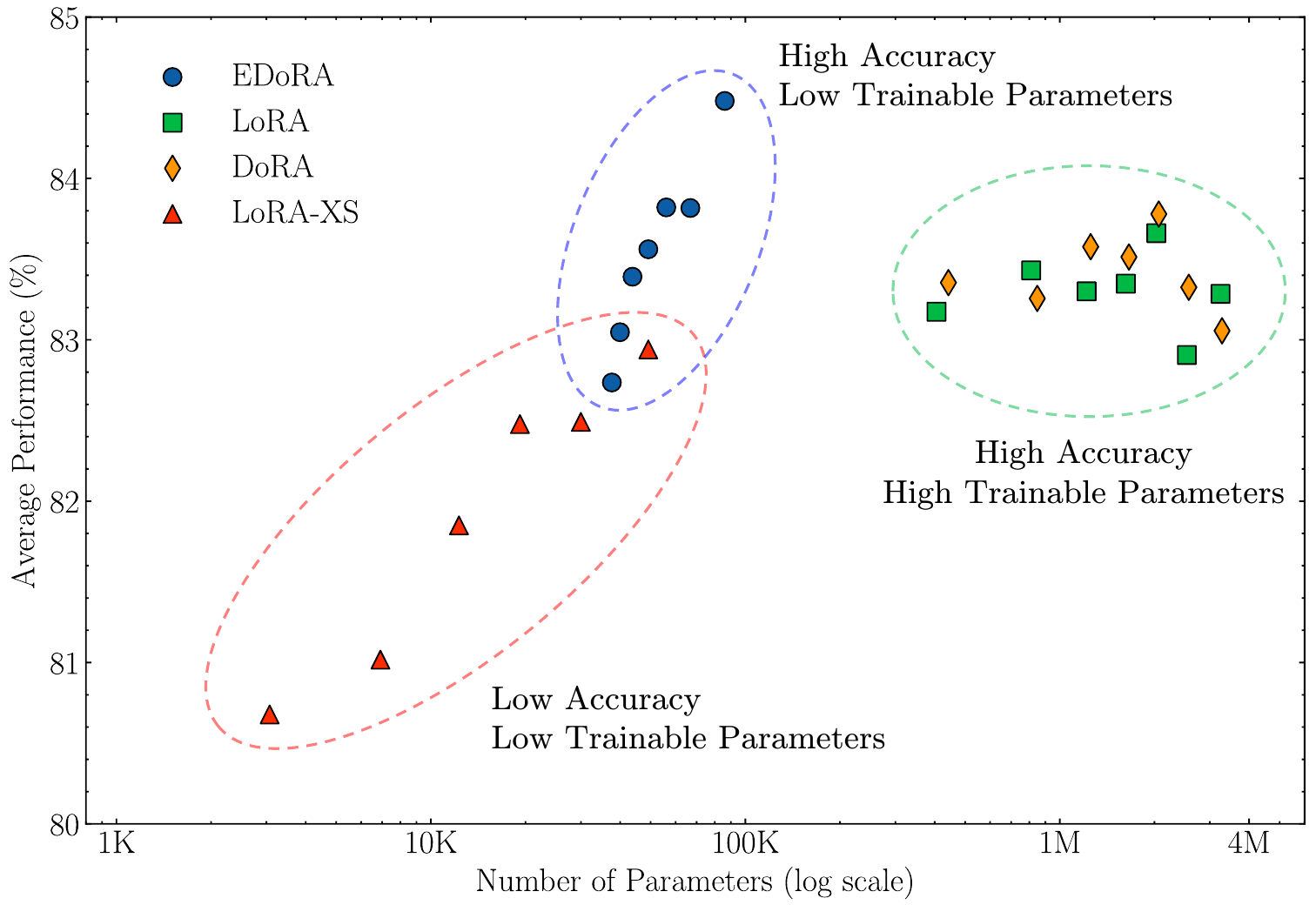}
    \caption{Relationship between the number of trainable parameters and average performance.}
    \label{fig:average_performance}
\end{figure}

\begin{figure}[t]
  \centering
    \includegraphics[width=0.475\textwidth]{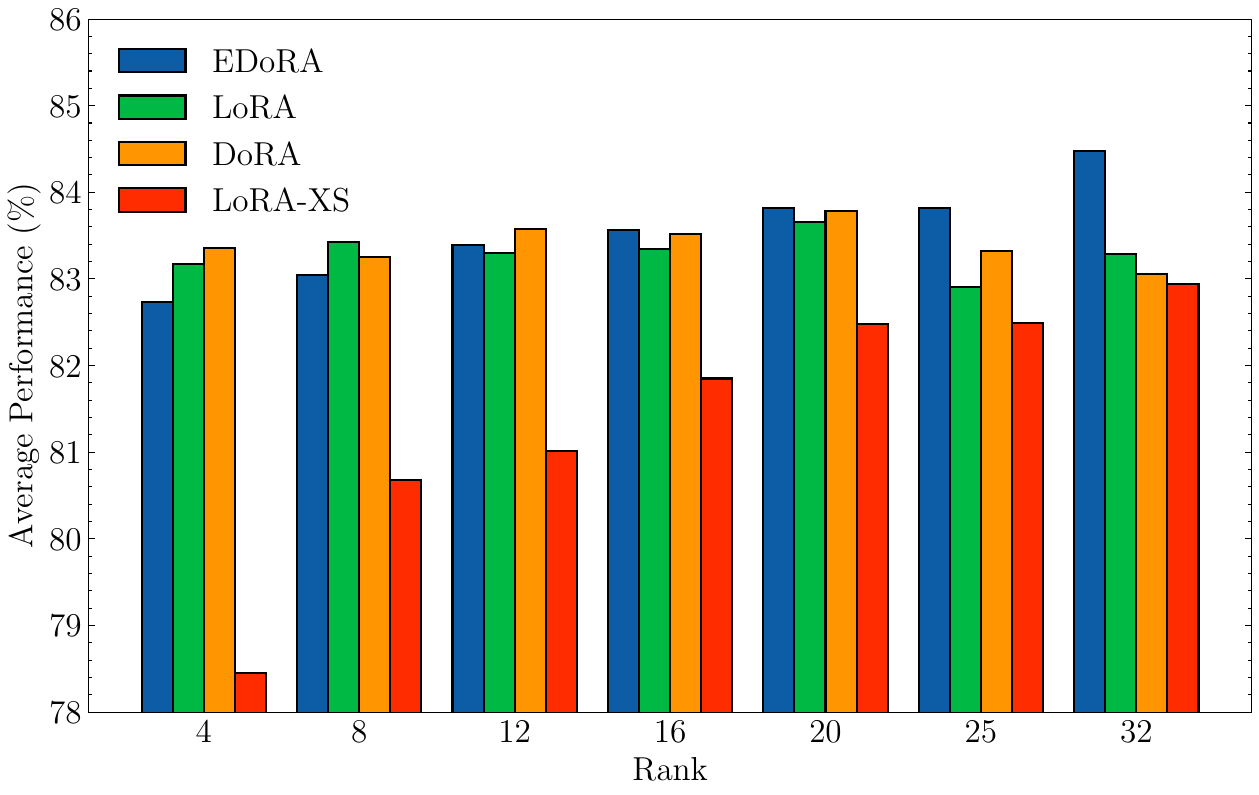}
    \caption{Impact of rank on average performance across different methods.}
    \label{fig:rank_average_performance}
\end{figure}

\begin{table*}[t]

\centering
\caption{Performance comparison of EDoRA with LoRA, DoRA and LoRA-XS on the GLUE benchmark. Higher is better for all columns except the last one; lower is better for this one.}
\label{tab:performance_glue}
\footnotesize
\begin{tabular}{l|c|c|p{13mm}p{12mm}p{12mm}p{12mm}p{12mm}p{14mm}|P{5mm}P{9mm}}
\toprule
Model & Rank & \begin{tabular}[c]{@{}c@{}}\#Trainable\\ Parameters\end{tabular} & RTE & STSB & MRPC & COLA & QNLI & SST2 & Avg. & \begin{tabular}[c]{@{}c@{}}\#Params\\(\%)\end{tabular} \\
\midrule
\multirow{7}{*}{DoRA} & 4 & 442K & $74.73_{\pm 1.06}$ & $90.50_{\pm 0.19}$ & $88.24_{\pm 0.97}$ & $60.38_{\pm 1.21}$ & $92.59_{\pm 0.11}$ & $93.69_{\pm 0.43}$ & \textbf{83.35} & 0.82 \\
 & 8 & 848K & $75.09_{\pm 2.14}$ & $90.63_{\pm 0.14}$ & $88.97_{\pm 0.48}$ & $58.36_{\pm 1.74}$ & $92.68_{\pm 0.06}$ & $93.81_{\pm 0.26}$ & 83.26 & 1.14 \\
 & 12 & 1.3M & $75.45_{\pm 1.31}$ & $90.65_{\pm 0.11}$ & $89.22_{\pm 1.17}$ & $59.65_{\pm 1.41}$ & $92.57_{\pm 0.12}$ & $93.92_{\pm 0.25}$ & \textbf{83.58} & 1.46\\
 & 16 & 1.7M & $75.45_{\pm 0.59}$ & $90.61_{\pm 0.12}$ & $88.97_{\pm 0.40}$ & $59.35_{\pm 2.34}$ & $92.66_{\pm 0.18}$ & $94.04_{\pm 0.31}$ & 83.51 & 1.77 \\
 & 20 & 2.1M & $76.17_{\pm 1.24}$ & $90.59_{\pm 0.18}$ & $88.73_{\pm 1.26}$ & $60.36_{\pm 1.13}$ & $92.79_{\pm 0.18}$ & $94.04_{\pm 0.29}$ & 83.78 & 2.09 \\
 & 25 & 2.6M & $75.45_{\pm 1.43}$ & $90.62_{\pm 0.13}$ & $88.48_{\pm 0.89}$ & $58.87_{\pm 1.07}$ & $92.49_{\pm 0.10}$ & $94.04_{\pm 0.44}$ & 83.33 & 2.48 \\
 & 32 & 3.3M & $73.65_{\pm 2.31}$ & $90.68_{\pm 0.13}$ & $88.24_{\pm 0.40}$ & $59.40_{\pm 1.25}$ & $92.57_{\pm 0.12}$ & $93.81_{\pm 0.22}$ & 83.06 & 3.01 \\
\midrule
\multirow{7}{*}{LoRA} & 4 & 405K & $74.73_{\pm 0.82}$ & $90.51_{\pm 0.14}$ & $87.75_{\pm 0.64}$ & $59.57_{\pm 0.92}$ & $92.57_{\pm 0.14}$ & $93.92_{\pm 0.44}$ & 83.17 & 0.79 \\
 & 8 & 811K & $75.09_{\pm 0.98}$ & $90.48_{\pm 0.11}$ & $89.22_{\pm 0.37}$ & $59.35_{\pm 0.63}$ & $92.64_{\pm 0.13}$ & $93.81_{\pm 0.31}$ & \textbf{83.43} & 1.11 \\
 & 12 & 1.2M & $74.37_{\pm 1.41}$ & $90.45_{\pm 0.12}$ & $88.73_{\pm 1.26}$ & $59.58_{\pm 0.36}$ & $92.64_{\pm 0.09}$ & $94.04_{\pm 0.29}$ & 83.30 & 1.43 \\
 & 16 & 1.6M & $73.65_{\pm 2.13}$ & $90.55_{\pm 0.16}$ & $88.48_{\pm 0.54}$ & $60.62_{\pm 0.85}$ & $92.64_{\pm 0.08}$ & $94.15_{\pm 0.40}$ & 83.35 & 1.75 \\
 & 20 & 2.0M & $74.73_{\pm 1.37}$ & $90.71_{\pm 0.22}$ & $\mathbf{89.46}_{\pm 1.08}$ & $60.36_{\pm 0.94}$ & $\mathbf{92.90}_{\pm 0.16}$ & $93.81_{\pm 0.40}$ & 83.66 & 2.06\\
 & 25 & 2.5M & $72.92_{\pm 1.84}$ & $90.56_{\pm 0.13}$ & $87.99_{\pm 1.12}$ & $59.58_{\pm 0.46}$ & $92.57_{\pm 0.14}$ & $93.81_{\pm 0.34}$ & 82.91 & 2.45 \\
 & 32 & 3.2M & $74.37_{\pm 1.16}$ & $90.65_{\pm 0.12}$ & $88.73_{\pm 0.40}$ & $59.10_{\pm 1.41}$ & $92.71_{\pm 0.18}$ & $94.15_{\pm 0.21}$ & 83.28 & 2.99\\
\hline
\rowcolor{gray!15}
 & 4 & 37.6K & $72.20_{\pm 0.75}$ & $90.73_{\pm 0.21}$ & $88.24_{\pm 1.39}$ & $60.36_{\pm 0.63}$ & $90.96_{\pm 0.18}$ & $93.92_{\pm 0.35}$ & 82.74 & 0.50 \\ \rowcolor{gray!15}
 & 8 & 39.9K & $72.20_{\pm 1.30}$ & $90.69_{\pm 0.10}$ & $87.25_{\pm 0.57}$ & $62.82_{\pm 0.95}$ & $91.16_{\pm 0.14}$ & $94.15_{\pm 0.20}$ & 83.05 & 0.50 \\
 \rowcolor{gray!15}
 & 12 & 43.7K & $74.37_{\pm 1.25}$ & $91.03_{\pm 0.16}$ & $87.25_{\pm 0.62}$ & $62.07_{\pm 1.04}$ & $91.36_{\pm 0.15}$ & $94.27_{\pm 0.37}$ & 83.39 & 0.50 \\
 \rowcolor{gray!15}
 & 16 & 49.2K & $74.73_{\pm 2.40}$ & $91.04_{\pm 0.22}$ & $87.50_{\pm 1.72}$ & $62.59_{\pm 1.51}$ & $91.36_{\pm 0.07}$ & $94.15_{\pm 0.29}$ & \textbf{83.56} & 0.51 \\
 \rowcolor{gray!15}
 & 20 & 56.0K & $75.09_{\pm 1.30}$ & $\mathbf{91.16}_{\pm 0.08}$ & $87.50_{\pm 0.88}$ & $\mathbf{63.06}_{\pm 0.28}$ & $91.73_{\pm 0.29}$ & $\mathbf{94.38}_{\pm 0.07}$ & \textbf{83.82} & 0.51 \\
 \rowcolor{gray!15}
 & 25 & 66.9K & $75.81_{\pm 1.50}$ & $91.07_{\pm 0.24}$ & $88.48_{\pm 1.23}$ & $61.82_{\pm 0.70}$ & $91.80_{\pm 0.09}$ & $93.92_{\pm 0.11}$ & \textbf{83.82} & 0.52 \\
 \rowcolor{gray!15}
 \multirow{-7}{*}{\shortstack{EDoRA\\(Ours)}}& 32 & 86.0K & $\mathbf{78.70}_{\pm 1.37}$ & $90.98_{\pm 0.14}$ & $88.97_{\pm 1.53}$ & $62.33_{\pm 1.08}$ & $91.87_{\pm 0.19}$ & $94.04_{\pm 0.13}$ & \textbf{84.48} & 0.53 \\

\hline

 & 4 & 0.8K & $67.15_{\pm 1.11}$ & $86.69_{\pm 0.80}$ & $86.27_{\pm 0.68}$ & $52.86_{\pm 1.30}$ & $85.98_{\pm 0.20}$ & $91.74_{\pm 0.32}$ & 78.45 & 0.47 \\
 & 8 & 3.1K & $70.40_{\pm 1.28}$ & $89.02_{\pm 0.19}$ & $86.76_{\pm 0.56}$ & $55.80_{\pm 0.84}$ & $88.85_{\pm 0.32}$ & $93.23_{\pm 0.08}$ & 80.68 & 0.47 \\
 & 12 & 6.9K & $69.31_{\pm 0.82}$ & $89.47_{\pm 0.24}$ & $87.50_{\pm 0.84}$ & $56.24_{\pm 1.08}$ & $90.01_{\pm 0.16}$ & $93.58_{\pm 0.31}$ & 81.02 & 0.47 \\
 & 16 & 12.3K & $71.84_{\pm 1.21}$ & $89.91_{\pm 0.19}$ & $86.76_{\pm 0.80}$ & $58.80_{\pm 0.61}$ & $90.32_{\pm 0.14}$ & $93.46_{\pm 0.31}$ & 81.85 & 0.48 \\
 & 20 & 19.2K & $72.92_{\pm 0.65}$ & $90.22_{\pm 0.11}$ & $87.50_{\pm 1.02}$ & $60.07_{\pm 0.74}$ & $90.81_{\pm 0.16}$ & $93.35_{\pm 0.27}$ & 82.48 & 0.48 \\
 & 25 & 30.0K & $72.20_{\pm 1.69}$ & $90.29_{\pm 0.12}$ & $87.25_{\pm 0.61}$ & $60.07_{\pm 1.02}$ & $91.32_{\pm 0.16}$ & $93.81_{\pm 0.32}$ & 82.49 & 0.49 \\
  \multirow{-7}{*}{LoRA-XS}& 32 & 49.2K & $74.01_{\pm 1.04}$ & $90.26_{\pm 0.13}$ & $87.25_{\pm 0.96}$ & $60.59_{\pm 0.33}$ & $91.49_{\pm 0.17}$ & $94.04_{\pm 0.17}$ & 82.94 & 0.50\\

\bottomrule
\end{tabular}
\end{table*}

We also observed that EDoRA demonstrates high effectiveness for single-sentence tasks (i.e., COLA and SST2), as it outperforms other methods. This is due to its ability to focus on efficient directional adaptation while retaining the core knowledge of the pre-trained model. EDoRA’s decomposition strategy allows it to replicate the fine-grained adjustments observed in full fine-tuning, which is particularly advantageous in single-sentence tasks. These tasks often require subtle changes in model parameters to interpret linguistic nuances or emotional tone. EDoRA's directional adaptation enables the model to capture these subtle changes effectively. Moreover, using SVD to initialize the low-rank matrices ensures that the adaptation starts from a subspace aligned with the most important features of the pre-trained model. This initialization likely aids in single-sentence tasks, where the model must recognize patterns closely related to linguistic correctness or sentiment without deviating significantly from the pre-trained model’s learned distributions.

Figure \ref{fig:average_performance} illustrates the relationship between the number of parameters and average performance. EDoRA obtained the highest average performance (84.48\%), demonstrating its robustness and adaptability in handling diverse tasks, and consistently outperformed LoRA-XS. When the number of trainable parameters in EDoRA and LoRA-XS are equal-rank 16 for EDoRA and rank 32 for LoRA-XS with 49.2K trainable parameters—EDoRA exceeds LoRA-XS’s performance by 0.62\%. The results highlight EDoRA’s ability to balance parameter efficiency and model performance effectively by making matrices $A$ and $B$ frozen, adding a small trainable matrix, and using SVD for the initialization. By drastically reducing the number of trainable parameters, EDoRA lowers storage costs, making it suitable for deployment on embedded devices and an efficient solution for large-scale applications such as adapting LLMs to downstream tasks where memory constraints are critical. As an intuitive example, using LoRA on the GPT-3 model \cite{brown2020language} with a rank of 16 while adapting only the query and value matrices requires 144MB of memory per checkpoint. Scaling this to serve 1 million personalized models would demand a total of 144TB of memory. Applying EDoRA in this scenario reduces the memory requirement to 4.59TB.

\begin{table*}[htbp]

\centering
\caption{The impact of initialization method on the EDoRA performance.}
\label{tab:perormance_SVD_Random}
\footnotesize
\begin{tabular}{c|c|P{15mm}P{15mm}P{15mm}P{15mm}P{15mm}P{15mm}P{5mm}}
\toprule
 Rank & \begin{tabular}[c]{@{}c@{}}Initialization\\Method\end{tabular}  & RTE & STSB & MRPC & COLA & QNLI & SST2 & Avg. \\

\midrule 
\multirow{2}{*}{4} & Random  & $70.40_{\pm 0.55}$ & $90.07_{\pm 0.32}$ & $87.25_{\pm 1.02}$ & $57.27_{\pm 0.70}$ & $90.68_{\pm 0.31}$ & $93.00_{\pm 0.34}$ & 81.45 \\
& SVD  & $72.20_{\pm 0.75}$ & $90.73_{\pm 0.21}$ & $88.24_{\pm 1.39}$ & $60.36_{\pm 0.63}$ & $90.96_{\pm 0.18}$ & $93.92_{\pm 0.35}$ & \textbf{82.74} \\
\midrule 
 \multirow{2}{*}{8} & Random  & $70.76_{\pm 1.57}$ & $90.33_{\pm 0.22}$ & $87.50_{\pm 0.88}$ & $56.50_{\pm 2.15}$ & $90.99_{\pm 0.30}$ & $93.23_{\pm 0.30}$ & 81.55 \\
  & SVD  & $72.20_{\pm 1.30}$ & $90.69_{\pm 0.10}$ & $87.25_{\pm 0.57}$ & $62.82_{\pm 0.95}$ & $91.16_{\pm 0.14}$ & $94.15_{\pm 0.20}$ & \textbf{83.05} \\
  \midrule 
 \multirow{2}{*}{12} & Random  & $72.20_{\pm 1.44}$ & $90.54_{\pm 0.09}$ & $87.75_{\pm 1.30}$ & $59.30_{\pm 1.48}$ & $90.90_{\pm 0.08}$ & $93.23_{\pm 0.30}$ & 82.32 \\
 & SVD  & $74.37_{\pm 1.25}$ & $91.03_{\pm 0.16}$ & $87.25_{\pm 0.62}$ & $62.07_{\pm 1.04}$ & $91.36_{\pm 0.15}$ & $94.27_{\pm 0.37}$ & \textbf{83.39} \\
 \midrule 
 \multirow{2}{*}{16} & Random  & $70.76_{\pm 1.16}$ & $90.52_{\pm 0.14}$ & $87.25_{\pm 0.37}$ & $56.77_{\pm 1.28}$ & $91.29_{\pm 0.17}$ & $93.46_{\pm 0.41}$ & 81.67 \\
& SVD  & $74.73_{\pm 2.40}$ & $91.04_{\pm 0.22}$ & $87.50_{\pm 1.72}$ & $62.59_{\pm 1.51}$ & $91.36_{\pm 0.07}$ & $94.15_{\pm 0.29}$ & \textbf{83.56} \\
\midrule 
 \multirow{2}{*}{20} & Random & $73.65_{\pm 1.50}$ & $90.63_{\pm 0.05}$ & $87.25_{\pm 1.36}$ & $58.29_{\pm 1.77}$ & $90.94_{\pm 0.25}$ & $93.81_{\pm 0.18}$ & 82.43 \\
 & SVD & $75.09_{\pm 1.30}$ & $\mathbf{91.16}_{\pm 0.08}$ & $87.50_{\pm 0.88}$ & $\mathbf{63.06}_{\pm 0.28}$ & $91.73_{\pm 0.29}$ & $\mathbf{94.38}_{\pm 0.07}$ & \textbf{83.82} \\
 \midrule 
 \multirow{2}{*}{25} & Random & $72.92_{\pm 0.72}$ & $90.72_{\pm 0.02}$ & $87.25_{\pm 0.37}$ & $59.31_{\pm 1.80}$ & $91.43_{\pm 0.22}$ & $93.46_{\pm 0.37}$ & 82.52 \\
 & SVD & $75.81_{\pm 1.50}$ & $91.07_{\pm 0.24}$ & $88.48_{\pm 1.23}$ & $61.82_{\pm 0.70}$ & $91.80_{\pm 0.09}$ & $93.92_{\pm 0.11}$ & \textbf{83.82}\\
 \midrule 
 \multirow{2}{*}{32} & Random & $75.45_{\pm 0.55}$ & $90.64_{\pm 0.10}$ & $86.27_{\pm 1.16}$ & $59.31_{\pm 1.60}$ & $91.43_{\pm 0.15}$ & $93.00_{\pm 0.30}$ & 82.68 \\
& SVD & $\mathbf{78.70}_{\pm 1.37}$ & $90.98_{\pm 0.14}$ & $\mathbf{88.97}_{\pm 1.53}$ & $62.33_{\pm 1.08}$ & $\mathbf{91.87}_{\pm 0.19}$ & $94.04_{\pm 0.13}$ & \textbf{84.48} \\

\bottomrule
\end{tabular}
\end{table*}

\begin{figure}[t]
  \centering
    \includegraphics[width=0.475\textwidth]{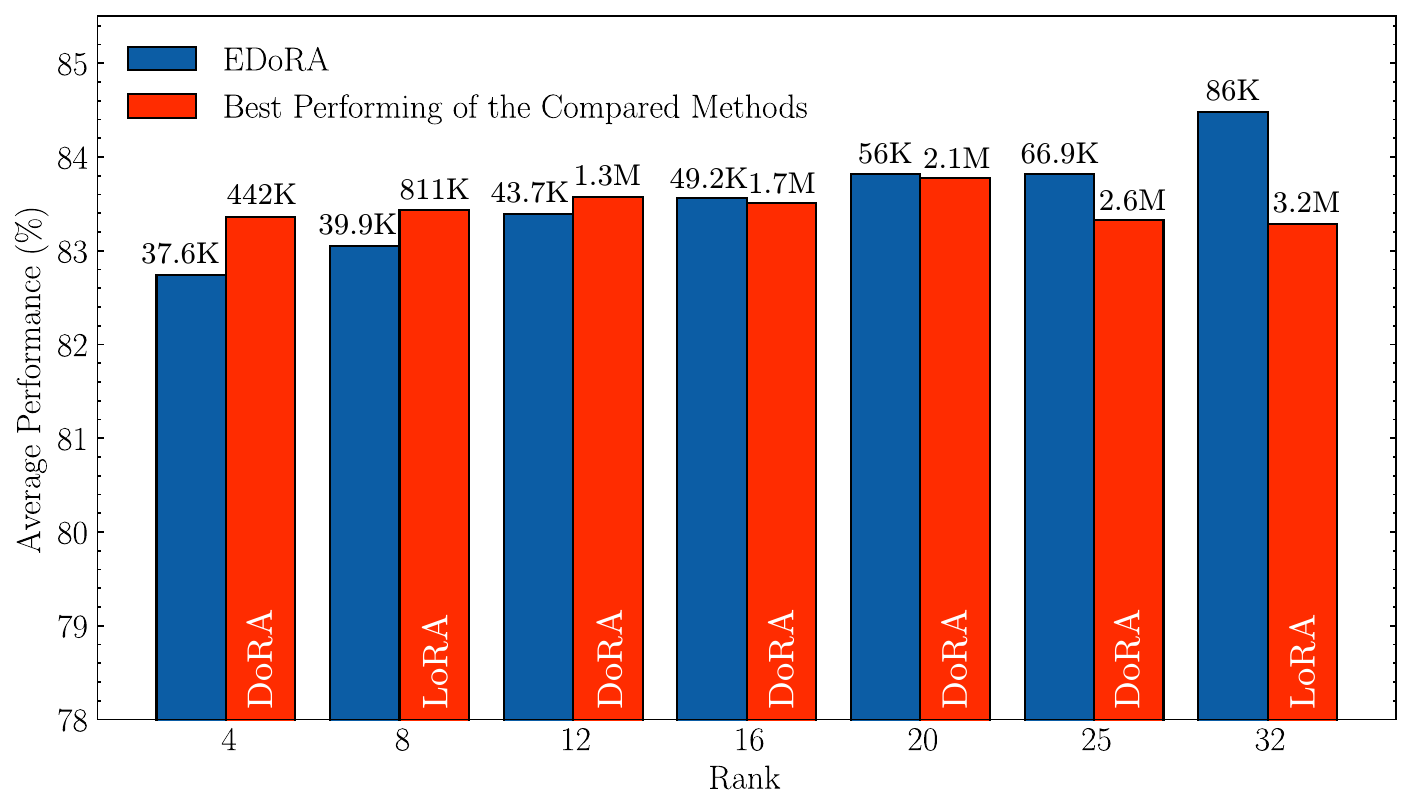}
    \caption{The impact of rank on average performance across different methods. The number of trainable parameters is shown above the bar plots for reference. At each rank, EDoRA is compared with LoRA, DoRA and LoRA-XS. For clarity, EDoRA’s average performance is displayed alongside the best-performing method among the three.}
    \label{fig:rank_average_performance_best}
\end{figure}

The average performance of various techniques for different ranks is shown in Figure \ref{fig:rank_average_performance}. It is evident that the performance of both EDoRA and LoRA-XS improves with increasing rank. These two methods have fewer parameters at lower ranks compared to LoRA and DoRA; as the rank increases, their trainable parameter count increases sufficiently to capture the knowledge in the training data effectively. EDoRA outperformed other methods from rank 16 to 32. As the rank increases, the difference in average performance between EDoRA and the second-best method also increases. Specifically, this difference is 0.05\% at rank 16 and increases to 1.2\% at rank 32. Figure \ref{fig:rank_average_performance_best} illustrates the impact of rank on the average performance of PEFT methods. It indicates that EDoRA consistently outperforms other methods with a lower parameter budget. At lower ranks (i.e., 4, 8 and 12), EDoRA achieved competitive performance while having a much smaller number of parameters. At higher ranks (i.e., 16, 20, 25 and 32), EDoRA outperformed other methods. For instance, at rank 16, EDoRA achieved the highest average performance despite having fewer trainable parameters (i.e., 49.2K vs. 1.7M). These results show that EDoRA can effectively leverage higher-rank matrices to capture more complex relationships in the data, leading to better adaptation.

\subsection{Ablation Study}

In this section, we conduct an ablation study to demonstrate the impact of SVD initialization on EDoRA's performance. We repeated the experiments using a random initialization instead of SVD initialization. The results, summarized in Table \ref{tab:perormance_SVD_Random}, indicate that SVD initialization consistently outperforms random initialization across all tasks and ranks. For instance, at rank 32, the average performance with SVD initialization is 84.48\%, compared to 82.68\% with random initialization. SVD initialization achieved, on average, 1.46\% better performance than the random initialization across different ranks. These results highlight the importance of initializing the low-rank matrices in a subspace aligned with the pre-trained weights' most important components, which SVD achieves effectively. In contrast, random initialization may start the adaptation process from a less optimal subspace, leading to slower convergence and reduced performance.

\section{Conclusion}
\label{conclusion}

This paper proposed EDoRA, a novel PEFT method that uses a decomposition strategy and SVD-based initialization to address the scalability and learning pattern challenges of existing approaches such as LoRA and DoRA. By decomposing pre-trained weights into magnitude and directional components, freezing low-rank matrices, and introducing a compact trainable matrix between them, EDoRA achieves substantial reductions in trainable parameters while maintaining high performance across diverse tasks. This reduction in parameters not only lowers computational and storage demands but also mitigates the risk of overfitting, especially in scenarios with limited training data. EDoRA's decomposition strategy aligns closely with the learning dynamics of full fine-tuning, further enhancing its adaptability and efficiency.  Experimental results on the GLUE benchmark demonstrated EDoRA’s effectiveness, achieving competitive or superior performance compared to state-of-the-art methods with up to 30x fewer trainable parameters. The results highlighted EDoRA's potential as a scalable and resource-efficient solution for adapting large language models to downstream tasks, making it particularly suitable for memory-constrained environments. Future work could explore extending EDoRA to multi-modal tasks, making rank of trainable matrices adaptive based on the importance of weights in each layer, and further optimizing its initialization strategies to enhance generalizability across even more complex domains.

\subsubsection*{Acknowledgments}
This work was supported by the Engineering and Physical Sciences Research Council (Fellowship
number EP/V007092/1).

\bibliography{references}
\bibliographystyle{icml2024}

\newpage
\appendix
\onecolumn
\section{Hyperparameter Configurations}

\begin{table}[!h]
    \centering
    \caption{Hyperparameter configuration of EDoRA for fine-tuning RoBERTa-base model on GLUE benchmark tasks.}
    \label{tab:hyperparameter}
 \begin{tabular}{@{}c|c|c|c|c@{}}
         \midrule
         Task & Rank & EDoRA Learning Rate & Classifier Learning Rate & Epochs \\
         \midrule

        \multirow{6}{*}{RTE}
        & 4 & 1E$-$3 & 5E$-$3 & \multirow{7}{*}{50} \\
        & 8 & 1E$-$3 & 5E$-$3 &  \\
        & 12 & 1E$-$3 & 6E$-$4 &  \\
        & 16 & 1E$-$3 & 5E$-$4 &  \\
        & 20 & 5E$-$3 & 1E$-$3 &  \\
        & 25 & 1E$-$3 & 6E$-$4 &  \\
        & 32 & 5E$-$3 & 6E$-$4 & \\
        \midrule

        \multirow{6}{*}{STSB}
        & 4 & 1E$-$3 & 1E$-$4 & \multirow{7}{*}{50} \\
        & 8 & 1E$-$3 & 1E$-$4 &  \\
        & 12 & 1E$-$3 & 5E$-$4 &  \\
        & 16 & 5E$-$3 & 5E$-$4 &  \\
        & 20 & 1E$-$3 & 1E$-$4 &  \\
        & 25 & 1E$-$3 & 1E$-$4 &  \\
        & 32 & 1E$-$3 & 1E$-$4 & \\
        \midrule

        \multirow{6}{*}{MRPC}
        & 4 & 5E$-$3 & 5E$-$3 & \multirow{7}{*}{50} \\
        & 8 & 5E$-$3 & 5E$-$3 &  \\
        & 12 & 5E$-$3 & 1E$-$4 &  \\
        & 16 & 5E$-$3 & 6E$-$4 &  \\
        & 20 & 5E$-$3 & 1E$-$4 &  \\
        & 25 & 5E$-$3 & 5E$-$3 &  \\
        & 32 & 5E$-$3 & 1E$-$4 & \\
        \midrule

        \multirow{6}{*}{COLA}
        & 4 & 1E$-$3 & 1E$-$4 & \multirow{7}{*}{50} \\
        & 8 & 1E$-$3 & 5E$-$3 &  \\
        & 12 & 1E$-$3 & 5E$-$3 &  \\
        & 16 & 6E$-$4 & 1E$-$3 &  \\
        & 20 & 1E$-$3 & 5E$-$3 &  \\
        & 25 & 1E$-$3 & 1E$-$3 &  \\
        & 32 & 1E$-$3 & 5E$-$4 & \\
        \midrule

        \multirow{6}{*}{QNLI}
        & 4 & 1E$-$3 & 5E$-$4 & \multirow{7}{*}{20} \\
        & 8 & 1E$-$3 & 6E$-$4 &  \\
        & 12 & 5E$-$4 & 6E$-$4 &  \\
        & 16 & 6E$-$4 & 6E$-$4 &  \\
        & 20 & 5E$-$4 & 6E$-$4 &  \\
        & 25 & 5E$-$4 & 6E$-$4 &  \\
        & 32 & 1E$-$3 & 5E$-$3 & \\
        \midrule

        \multirow{6}{*}{SST2}
        & 4 & 5E$-$4 & 5E$-$3 & \multirow{7}{*}{20} \\
        & 8 & 6E$-$4 & 1E$-$3 &  \\
        & 12 & 6E$-$4 & 1E$-$3 &  \\
        & 16 & 6E$-$4 & 5E$-$4 &  \\
        & 20 & 5E$-$4 & 1E$-$4 &  \\
        & 25 & 5E$-$4 & 1E$-$4 &  \\
        & 32 & 5E$-$4 & 5E$-$4 & \\
         \bottomrule
    \end{tabular}

    \label{tab:gluehyperparams}
\end{table}


\end{document}